\documentclass[conference]{IEEEtran}
\IEEEoverridecommandlockouts
\usepackage{cite}
\usepackage{amsmath,amssymb,amsfonts}
\usepackage{algorithmic}
\usepackage{graphicx}
\usepackage{textcomp}
\usepackage{xcolor}
\def\BibTeX{{\rm B\kern-.05em{\sc i\kern-.025em b}\kern-.08em
    T\kern-.1667em\lower.7ex\hbox{E}\kern-.125emX}}

\newcommand{\RR}[1]{\mathbb{R}^{#1}} 
\newcommand{\bd}[1]{\mbox{\boldmath $#1$}}
\newcommand{\entrconst}{2 \pi e}
\newcommand{\MVImu}{MVI\textsubscript{$\mu$}}
\newcommand{\MVIlr}{MVI\textsubscript{lr}}
\newcommand{\MVIeig}{MVI\textsubscript{eig}}
\newcommand{\VIdiag}{VI\textsubscript{diag}}

\DeclareMathOperator{\diag}{diag}
\DeclareMathOperator{\DKL}{D\textsubscript{KL}} % if changes occur here, then update abstract manually

\begin{document}
%####################################################################################################
%####################################################################################################
%####################################################################################################

\title{Mixed Variational Inference\\
\thanks{The author gratefully acknowledges the generous and invaluable support of the Klaus Tschira Foundation. }
}

\author{\IEEEauthorblockN{Nikolaos Gianniotis}
\IEEEauthorblockA{\textit{Astroinformatics} \\
\textit{Heidelberg Institute for Theoretical Studies gGmbH}\\
Heidelberg, Germany \\
nikos.gianniotis@h-its.org}
}

\maketitle

\begin{abstract}
The Laplace approximation has been one of the workhorses of Bayesian inference.
It often delivers  good approximations in practice despite the fact that 
it does not strictly take into account where the volume of posterior density lies.
Variational approaches avoid this issue by explicitly minimising the Kullback-Leibler divergence D\textsubscript{KL} between
a postulated posterior and the true (unnormalised) logarithmic posterior.
However,  they rely on a closed form D\textsubscript{KL} in order to
update  the variational parameters. To address this, stochastic versions of variational inference
have been devised that approximate the intractable D\textsubscript{KL} with a Monte Carlo average. This approximation allows calculating
gradients with respect to the variational parameters. However, variational methods often postulate a factorised Gaussian approximating posterior.
%in order to limit the number of free variational parameters that need to be optimised. 
In doing so, they sacrifice a-posteriori correlations.
In this work, we propose a  method that combines the Laplace approximation with the variational approach.
The advantages are that we maintain: applicability on non-conjugate models, posterior correlations and a reduced number of free variational parameters.
Numerical experiments demonstrate improvement over the Laplace approximation and variational inference with factorised Gaussian posteriors.
\end{abstract}

\begin{IEEEkeywords}
Bayesian inference,  variational techniques, Laplace approximation, mean field, Gaussian approximation%, approximation methods
\end{IEEEkeywords}

%#####################################################################################################
\section{Introduction}
%#####################################################################################################

Bayesian inference provides a way of making use of the complete information  available either as data or prior knowledge.
It enables us to capture the uncertainty present in the data and model assumptions, and propagate it to further tasks such as prediction and decision making.
However, exact Bayesian inference is only possible whenever mathematically convenient priors are combined with particular likelihood functions (e.g.~conjugacy).
Deviation from such convenience, results in intractable calculations that call for approximations.
%, unless one is willing to resort to Monte Carlo sampling methods.
%These are typically computationally very intensive but capable of delivering very accurate empirical approximations to the sought posterior densities without the need of assumming
% a constrained functional form for the approximating posterior.

The Laplace approximation (LA) has helped the advancement of Bayesian methodology in the machine learning field \cite{Mackay1992}
and has also been an important tool for practitioners \cite{Sivia2006}. LA produces a Gaussian approximating posterior.
In doing so, it operates myopically  in the sense that it determines the mean and covariance simply
by looking locally around the mode instead of focusing on where the volume of the density actually lies.
Despite this shortcoming, it has been found to produce good  approximations in a variety of contexts.
In a Gaussian process binary classification setting \cite{Nickisch2008}, LA is found to be on a par with other approximations in terms of error rate, though it performed poorer on other criteria.
The work in \cite{eskin2004laplace} employs LA in order to calculate approximate intractable integrals within the Expectation Propagation algorithm \cite{minka2001expectation}.
In \cite{Jensen2013} LA is found to perform well when compared to Expectation Propagation in a bounded regression task.
In \cite{Wu2017} LA is used to formulate an approximate form of the marginal likelihood that facilitates the update of hyperparameters
without the need to update the current Gaussian posterior given by LA.

Variational inference has been put forward (VI) \cite{Beal2003, Tzikas2008} as a solution to calculating posterior distributions in situations where certain expectations are not analytically tractable.
Its use has been widespread in eliciting posterior densities in e.g.~dimensionality reduction \cite{soh2016distance}, classification \cite{kaban2007bayesian}, regression \cite{Tzikas2008}, density estimation \cite{Yu2006} and in specialised applications like in astronomy \cite{regier2015celeste}.
%VI typically constrains the approximating posterior to a particular density form and then minimises the Kullback-Leibler divergence ( \DKL \) between the approximating and the true (unnormalised) posterior. 
The applicability of VI depends on choosing a posterior density form that allows the Kullback-Leibler divergence ($\DKL$) between the approximating and the true (unnormalised) posterior to be calculated in closed form. However, such a choice may not always exist.

Whenever it is not possible to obtain a closed-form  $\DKL$ within the VI framework, approximations become necessary.
One type of approximation approximates the logarithmic (unnormalised) posterior.
In \cite{Chappell2009} the logarithmic posterior is linearised via a first-order Taylor expansion which allows then calculating the expectation with respect to the approximating posterior in the  $\DKL$.
In a similar vein, second order Taylor expansions are considered in \cite{Woolrich2006}. 
Interestingly, \cite{Gershman2012} considers multiple second order Taylor expansions at different parameter locations of the logarithmic posterior.
This results in an approximate posterior density expressed as a mixture of spherical Gaussians that has the potential to capture multiple modes.
% Therein, the entropy term in the associated variational lower bound ensures that the individual components in the mixture automatically repel each other.
A second type of approximation  \cite{paisley2012variational, Salimans2013, Titsias2014doubly, kingma2013auto}  approximates the  $\DKL$ as a Monte Carlo average with samples drawn from the approximating posterior.
The resulting expression allows calculating the gradient with respect to the free variational parameters.
An update of the variational parameters follows, typically using a small step size in a stochastic gradient descent setting, after which the  $\DKL$ is approximated with a new Monte Carlo average.
In a slightly manner, \cite{Gianniotis2016, depraetere2017comparison} fix the Monte Carlo average approximation of the  $\DKL$ throughout the optimisation of the variational parameters. We finally note that often in practice e.g. \cite{paisley2012variational, Titsias2014doubly}, VI methods choose to work with a factorised Gaussian posterior which
has the advantage of reducing the number of free variational parameters that need to be optimised, but  also has the inevitable disadvantage of discarding potential parameter correlations in the posterior.
% discards potential correlations, e.g. \cite{paisley2012variational, Titsias2014doubly} while reducing the number of free variational parameters that need to be optimised.

In this work, we propose a method that combines the Laplace approximation with the variational approximation.
The method works on non-conjugate models, captures a-posteriori correlations and limits the number of free variational parameters.
The main idea is to take the Gaussian posterior obtained from the Laplace approximation, plug it into the variational lower bound and adapt it by optimising the lower bound.
The crux of the approach is to allow only a partial update of the Laplace Gaussian posterior.

% linearization results in an algorithm that is no longer ?pure? VB, the main consequence of this is that the guarantee of conver- gence for VB no longer applies

%#####################################################################################################
\section{Approximate Bayesian Inference}
%#####################################################################################################

We briefly review methods for approximate inference as a gentle reminder and for the purpose of introducing relevant notation.
We write the log-posterior as the sum of the model log-likelihood,  log-prior and minus log-evidence: 
\begin{equation}
\ln p(\bd{w}|\bd{\theta},\mathcal{D}) = \ln p(\mathcal{D} | \bd{w}, \bd{\theta}_\ell) + \ln p(\bd{w} | \bd{\theta}_\pi) - \ln\mathcal{Z} \ ,
\end{equation}
where $\mathcal{D}$ are the data, $\bd{w}\in\RR{D}$ are the model parameters and $\mathcal{Z} = \int  p(\mathcal{D} | \bd{w}, \bd{\theta}_\ell)  p(\bd{w} | \bd{\theta}_\pi) \bd{dw}$.
The log-likelihood and log-prior terms have hyperparameters $\bd{\theta}_\ell$ and $ \bd{\theta}_\pi$ which are  jointly summarised as  \bd{\theta} in the log-posterior.
In the following, the evidence is a constant which we discard.
Discarding it, gives us the unnormalised log-posterior  $\ln \tilde{p}(\bd{w}| \bd{\theta}, \mathcal{D})$.

%------------------------------------------------------------------------------------------------------------------------
\subsection{Laplace approximation}
%------------------------------------------------------------------------------------------------------------------------

The Laplace approximation (LA) seeks the mode\footnote{Multiple modes may be present.} $\bd{w}^{*}$ of the log-posterior density $\ln \tilde{p}(\bd{w}|\bd{\theta}, \mathcal{D})$
where $\bd{0} = \nabla_{\tiny \bd{w}} \ln  \tilde{p}(\bd{w}|\bd{\theta}, \mathcal{D})  \vert_{\scriptstyle \bd{\scriptstyle w} = \bd{\scriptstyle w}^{*}}  $. 
This may be carried out with gradient-based optimisation. % e.g.~using scaled-conjugate gradients.
At the found mode, we calculate the Hessian matrix $\bd{H} = \nabla \nabla  \ln \tilde{p}(\bd{w}|\bd{\theta}, \mathcal{D})  \vert_{\scriptstyle \bd{\scriptstyle w} = \bd{\scriptstyle w}^{*}} $.
The obtained approximating Gaussian posterior reads:
\begin{equation}
q(\bd{w}) = \mathcal{N}(\bd{w} | \bd{\mu}_{LA} = \bd{w}^{*}, \bd{\Sigma}_{LA} = -\bd{H}^{-1}) \ .
\end{equation}
We see that the covariance of the approximating posterior $q(\bd{w})$ is given by the local curvature of the posterior at the found mode.
The approximation can be good, if the true posterior concentrates strongly around the mode.

%------------------------------------------------------------------------------------------------------------------------
\subsection{Variational Inference}
%------------------------------------------------------------------------------------------------------------------------

Variational inference (VI)  \cite{Beal2003}  postulates an approximating posterior $q(\bd{w})$.
VI finds the $q(\bd{w})$
that maximises the following  $\DKL$ based objective, also known as the variational lower bound  \cite[Chapter $10$]{Bishop} :
\begin{align}
- \DKL (q(\bd{w}) || \tilde{p}(\bd{w}|\bd{\theta}, \mathcal{D})) & =  \int q(\bd{w}) \ln  \tilde{p}(\bd{w}|\bd{\theta}, \mathcal{D}) \bd{dw} \notag \\
                                                                              &\ \ \ \  - \int q(\bd{w}) \ln q(\bd{w}) \bd{dw} \ .
\label{eq:objective_VI}                                                                              
\end{align}
In general, VI does not require that $q(\bd{w})$ is a Gaussian, but here we choose to
work with  $q(\bd{w})=\mathcal{N}(\bd{w}|\bd{\mu},\bd{\Sigma})$. For this choice, the above objective now reads as:
\begin{equation}
\int \mathcal{N}(\bd{w}|\bd{\mu},\bd{\Sigma}) \ln  \tilde{p}(\bd{w}|\bd{\theta}, \mathcal{D}) \bd{dw} + \frac{1}{2}\ln |\entrconst \bd{\Sigma}| \ ,
\label{eq:objective_VI_gaussian}
\end{equation}
where the second term is the Gaussian entropy.
The free parameters in objective \eqref{eq:objective_VI_gaussian} are the variational parameters \bd{\mu}, \bd{\Sigma}
and hyperparameters  \bd{\theta}.

%------------------------------------------------------------------------------------------------------------------------
\subsection{Stochastic variational inference}
%------------------------------------------------------------------------------------------------------------------------

Stochastic variational inference \cite{Titsias2014doubly} addresses the difficulty that arises when
the expectation in the first term of \eqref{eq:objective_VI_gaussian}  is not tractable.
It does so by approximating the expectation by a Monte Carlo average with samples drawn from 
$\bd{w}_s \sim q(\bd{w})$:
\begin{equation}
\frac{1}{S} \sum_{s=1}^S \ln  \tilde{p}(\bd{w}_s | \bd{\theta}, \mathcal{D}) + \frac{1}{2} \ln |\entrconst \bd{\Sigma}| \ .
\label{eq:objective_SVI_q}
\end{equation}
The variational parameters no longer appear in the approximation, but it is possible to reintroduce them using the reparametrisation
$\bd{w}_s = \bd{\mu} + \bd{C} \bd{z}_s$, in terms of samples $\bd{z}_s \sim \mathcal{N}(\bd{0},\bd{I}_D)$:
\begin{equation}
\frac{1}{S} \sum_{s=1}^S \ln  \tilde{p}(\bd{\mu}+\bd{C}\bd{z}_s |\bd{\theta}, \mathcal{D}) + \frac{1}{2} \ln|\entrconst \bd{\Sigma}| \ ,
\label{eq:objective_SVI}
\end{equation}
where \bd{C} is a matrix\footnote{A common choice is the Cholesky decomposition.} such that $\bd{C}\bd{C}^T =  \bd{\Sigma}$.  
The free parameters in objective \eqref{eq:objective_SVI} are \bd{\mu}, \bd{\Sigma} and  \bd{\theta}.
We note that, typically, one chooses covariance \bd{\Sigma} to be a diagonal matrix  (e.g. \cite{paisley2012variational, Titsias2014doubly})
in order to limit the number of free variational parameters to be optimised. In this case, $q(\bd{w})$ is a factorised posterior.

%#####################################################################################################
\section{Proposed method}
%#####################################################################################################

The motivation behind this work is to apply VI on non-conjugate models using an approximating posterior $q(\bd{w})$ that captures a-posteriori 
correlations but at the same time limits the number of free variational parameters that need to be optimised.
To that end, we make use of the covariance $\bd{\Sigma}_{LA}$ of the approximating posterior $q(\bd{w})$ obtained via LA
and the approximate variational lower bound in \eqref{eq:objective_SVI}.
Since the proposed method combines  LA with the variational lower bound, we name it \textit{mixed variational inference} (MVI).
In the following,  we propose three ways that MVI can exploit the correlation structure present in $\bd{\Sigma}_{LA}$.

%-------------------------------------------------------------------------------------------------------------------
\subsection{Adaptation of mean only - \MVImu}
\label{sec:mvi_mu}
%-------------------------------------------------------------------------------------------------------------------

We perform the following Cholesky decomposition:
\begin{equation}
\bd{\Sigma}_{LA} = \bd{C}_{LA} \bd{C}_{LA}^T \ .
\end{equation}
We propose the posterior $q(\bd{w})=\mathcal{N}(\bd{w}| \bd{\mu},\bd{\Sigma}_{LA})$ and use it in the
approximate variational lower bound in \eqref{eq:objective_SVI}, which results in the following objective:
\begin{equation}
\frac{1}{S} \sum_{s=1}^S \ln \tilde{p}(\bd{\mu} + \bd{C}_{LA} \bd{z}_s |\bd{\theta}, \mathcal{D}) + \frac{1}{2} \ln|\entrconst \bd{\Sigma}_{LA}| \ ,
\label{eq:objective_mvi_mu}
\end{equation}
The free parameters in \eqref{eq:objective_mvi_mu} are the mean \bd{\mu} and  hyperparameters \bd{\theta}. % which is initialised to $\bd{\mu}_{LA}$ at the beginning of the optimisation.
Effectively, the proposed posterior is the Laplace posterior with the added flexibility of shifting its mean while keeping its covariance fixed to $\bd{\Sigma}_{LA}$.
Note, that here the entropy is a constant term that can be discarded during optimisation.

%-------------------------------------------------------------------------------------------------------------------
\subsection{Mean and scaling of covariance - \MVIeig}
\label{sec:mvi_eig}
%-------------------------------------------------------------------------------------------------------------------

We perform the following eigenvalue decomposition:
\begin{equation}
\bd{\Sigma}_{LA} = \bd{Q}_{LA}\diag(\bd{r}_{LA}^2)\bd{Q}_{LA}^T \ , 
\end{equation}
where matrix $\bd{Q}_{LA}\in\RR{D\times D}$ and vector $\bd{r}_{LA}\in\RR{D}$ hold the eigenvectors and square roots of the eigenvalues 
respectively\footnote{Operator $\diag$ creates a diagonal matrix using the vector it is applied to. Notation $\bd{r}^2$ implies raising the components of vector $\bd{r}$ to the power of $2$.}.
We propose $q(\bd{w})=\mathcal{N}(\bd{w}|\bd{\mu}, \bd{Q}_{LA}\diag(\bd{r}^2)\bd{Q}_{LA}^T)$ and optimise:
\begin{equation}
\frac{1}{S} \sum_{s=1}^S \ln \tilde{p}(\bd{\mu} +  \bd{Q}_{LA}\diag(\bd{r}) \bd{z}_s | \bd{\theta}, \mathcal{D} ) + \frac{1}{2} \ln|\entrconst \diag(\bd{r}^2)| \ .
\label{eq:objective_mvi_eig}
\end{equation}
The free parameters in \eqref{eq:objective_mvi_eig} are \bd{\mu}, \bd{r} and \bd{\theta}.
Note the simplification in the entropy term due to the orthogonal $\bd{Q}_{LA}$, i.e.~$|\entrconst   \diag(\bd{r}^2) \bd{Q}_{LA}^T \bd{Q}_{LA}| = |\entrconst \diag(\bd{r}^2)|$.
Effectively, the proposed posterior is the Laplace posterior which now has the added flexibility to shift the mean and scale the covariance matrix along its axes by adapting vector $\bd{r}$.

%-------------------------------------------------------------------------------------------------------------------
\subsection{Mean and low rank update of covariance - \MVIlr}
\label{sec:mvi_lr}
%-------------------------------------------------------------------------------------------------------------------

We introduce the vectors $\bd{U}, \bd{V} \in \RR{D}$.
We use the Cholesky decomposition $\bd{\Sigma}_{LA} = \bd{C}_{LA} \bd{C}_{LA}^T$
and form the matrix $\bd{L} = \bd{C}_{LA} + \bd{U}\bd{V}^T$.
We propose the posterior $q(\bd{w})=\mathcal{N}(\bd{w}|\bd{\mu}, \bd{L}\bd{L}^T)$ and the 
associated objective:
\begin{equation}
\frac{1}{S} \sum_{s=1}^S \ln \tilde{p}(\bd{\mu} +  \bd{L}\bd{z}_s | \bd{\theta}, \mathcal{D} ) + \frac{1}{2} \ln|\entrconst \bd{L}\bd{L}^T| \ .
\label{eq:objective_mvi_lr}
\end{equation}
The free parameters in \eqref{eq:objective_mvi_lr} are \bd{\mu}, \bd{U}, \bd{V} and \bd{\theta}.
Effectively, the proposed posterior is the Laplace posterior which now has the added flexibility to shift the mean but also modify its covariance matrix via a low-rank update.

The proposed posteriors are summarised in Table \ref{tbl:mvi_summary}.

%-------------------------------------------------------------------------------------------------------------------
\subsection{Initialisation}
\label{sec:initialisation}
%-------------------------------------------------------------------------------------------------------------------

We use the mean $\bd{\mu}_{LA}$ and optimised hyperparameters $\bd{\theta}_{LA}$ obtained from the Laplace approximation to 
initialise the mean in $q(\bd{w})=\mathcal{N}(\bd{w}|\bd{\mu}=\bd{\mu}_{LA},\bd{\Sigma})$ and hyperparameters $\bd{\theta}=\bd{\theta}_{LA}$
in each of the three proposed objectives.
We emphasize that the covariance in \MVImu\ is initialised to $\bd{\Sigma}_{LA}$ and fixed.
Vector \bd{r} in \MVIeig\ is initialised to the square root of the eigenvalues $\bd{r}_{LA}$.
Vectors \bd{U},\bd{V} in \MVIlr\ are randomly initialised by drawing them from $\mathcal{N}(\bd{0}, 0.01\bd{I}_D)$.

%-------------------------------------------------------------------------------------------------------------------
\subsection{Optimisation}
%-------------------------------------------------------------------------------------------------------------------

 \begin{table}[!t]
\renewcommand{\arraystretch}{1.3}
\caption{Summary of MVI posteriors. Variables with the subscript LA are fixed parameters (not optimised)
whose values are given by either the Cholesky or eigenvalue decomposition.}
\label{table_example}
\centering
\begin{tabular}{c|c|c|c}
\hline
\bfseries                   & \bfseries \MVImu                       & \bfseries \MVIeig & \bfseries \MVIlr   \\
\hline\hline
\#  parameters         &            D                                    &            2D            & 3D                       \\
mean                       &          \bd{\mu}                           &         \bd{\mu}       &          \bd{\mu}     \\
covariance  ``root''      &  $\bd{C}_{LA}$   &     $\bd{Q}_{LA}\diag(\bd{r})$    &      $\bd{C}_{LA} + \bd{U}\bd{V}^T$     \\
\hline
\end{tabular}
\label{tbl:mvi_summary}
\end{table}

Following \cite{Gianniotis2016, depraetere2017comparison} we draw $S$ number of samples $\bd{z}_s \sim \mathcal{N}(\bd{0},\bd{I}_D)$
which we keep fixed throughout the optimisation of the objectives in \eqref{eq:objective_mvi_mu}, \eqref{eq:objective_mvi_eig} and \eqref{eq:objective_mvi_lr}. This enables the use of 
scaled-conjugate gradients (SCG) as the optimisation routine\cite{mogensen2018optim} in contrast to the typically employed
stochastic gradient descent\footnote{We note that in \cite{Titsias2014doubly} the use of stochastic gradient is additionally motivated 
by the desire to train with ``mini-batches''.} \cite{Titsias2014doubly}. We note that the proposed method can in principle also employ the same optimisation scheme as in \cite{Titsias2014doubly}.
The free parameters \bd{\mu}, \bd{\theta} and the ones pertaining to the covariance in each proposed posterior are jointly optimised via SCG.
In all experiments we fix the number of drawn samples $\bd{z}_s$ to $S=10^3$.

%#####################################################################################################
\section{Numerical setup}
%#####################################################################################################

%-------------------------------------------------------------------------------------------------------------------
\subsection{Comparisons}
%-------------------------------------------------------------------------------------------------------------------

The proposed work builds on LA and VI in order to improve the performance (see section \ref{sec:measuring_performance}) of the Laplace approximation
and do better than VI when employing a factorised Gaussian posterior.
Specifically, the proposed posterior for the latter reads $\bd{q}(\bd{w}) = \mathcal{N}(\bd{w} | \bd{\mu}, \diag(\bd{\sigma}^2))$ and has a diagonal covariance
matrix whose elements are specified by the vector $\bd{\sigma}\in\RR{D}$. The associated objective reads:
\begin{equation}
\frac{1}{S} \sum_{s=1}^S \ln \tilde{p}(\bd{\mu} +  \diag(\bd{\sigma})\bd{z}_s | \bd{\theta}, \mathcal{D} ) + \frac{1}{2} \ln|\entrconst \diag(\bd{\sigma}^2)| \ .
\label{eq:objective_vi_diag}
\end{equation}
The free parameters in \eqref{eq:objective_vi_diag} are \bd{\mu}, \bd{\sigma} and \bd{\theta}.
We refer to this method as \VIdiag.

In the numerical experiments, we initialise the mean and hyperparameters with $\bd{\mu}=\bd{\mu}_{LA}$ and $\bd{\theta}=\bd{\theta}_{LA}$.
Regarding $\bd{\sigma}^2$, we experimented with two initialisations: either setting the elements of $\bd{\sigma}^2$ equal to the diagonal elements of $\bd{\Sigma}_{LA}$, or 
all equal to $10^{-4}$. In the experiments of Section \ref{sec:applications} we report for \VIdiag\ the best performance achieved by either initialisation.

%-------------------------------------------------------------------------------------------------------------------
\subsection{Measuring performance}
\label{sec:measuring_performance}
%-------------------------------------------------------------------------------------------------------------------

In the experiments of Section \ref{sec:applications}, we measure performance in terms of the logarithmic predictive density (LPD) (i.e.~marginal log-likelihood)
evaluated on test data:
\begin{align}
\ln p(\mathcal{D}_{test} | \bd{\theta})  &=               \ln \int p(\mathcal{D}_{test} | \bd{w}, \bd{\theta}) q(\bd{w}) \bd{dw} \notag \\
                                                            &\approx    \ln \frac{1}{S^\prime} \sum_{s^\prime=1}^{S^\prime}  p(\mathcal{D}_{test} | \bd{w}_{s^\prime}, \bd{\theta}) \ .
\label{eq:mll}
\end{align}
The LPD is approximated by $S^\prime$ number of samples drawn from $\bd{w}_{s^\prime} \sim\bd{q}(\bd{w})$, where $\bd{q}(\bd{w})$ is
the respective posterior obtained via LA, MVI or \VIdiag.
In all numerical experiments we use $S^\prime = 10^4$.

Along LPD, we also report error rates. 
For the regression problem  we report the mean squared error (MSE).
For the classification problems, the error rate is the
percentage of predicted labels not matching the true labels.

We compare MVI to LA and \VIdiag\ on a number of datasets as detailed in the corresponding sections.
Each dataset is split $100$ times into a training and testing set. 
The algorithms are run on each split, hence, we collect $100$ samples of the
algorithms' performance in terms of LPD and error rate on the test set. For each  dataset, we report the median LPD and median error rate on the test set for each algorithm. The best performance is marked with bold in the tables reporting the results.

Moreover, we attempt to detect whether the observed differences in median, over the $100$ collected performances, are statistically significant.
In the experiments, we observed that the collected performances are not normally distributed which precludes the use of a paired T-test. 
The Wilcoxon signed rank test is also precluded as it requires \cite[Chapter 4.7]{dalgaard2008introductory} that the distribution of the difference in median of the tested pairs is symmetric. Therefore, we resort to using a sign test  \cite[Chapter 2.5.2]{dalgaard2008introductory}, to check whether a difference in median performance exists (i.e. better or worse, but not by how much),  and
the confidence intervals constructed by the bootstrap \cite{davison1997bootstrap}.

We test whether the performance of the best algorithm (marked in the tables with bold)  is statistically significantly better by checking two conditions:
we pair the best performing algorithm with all other algorithms and carry out the sign test. The first condition is satisfied if for \textit{each pair}, the sign test rejects the null hypothesis that the median of the best algorithm is equal to the median of its respective paired algorithm. The second condition is satisfied if the $95\%$ confidence interval constructed by the bootstrap on the difference of the paired medians does not contain the $0$ value.
That is, if the $95\%$  confidence interval does not contain $0$, then $0$ is not a likely value for the difference in the true medians.
If both conditions are satisfied, we declare the best performance as statistically significant and mark it with a $\bullet$ marker in the respective tables.

\begin{figure}[t!]
\centerline{\includegraphics[scale=0.61]{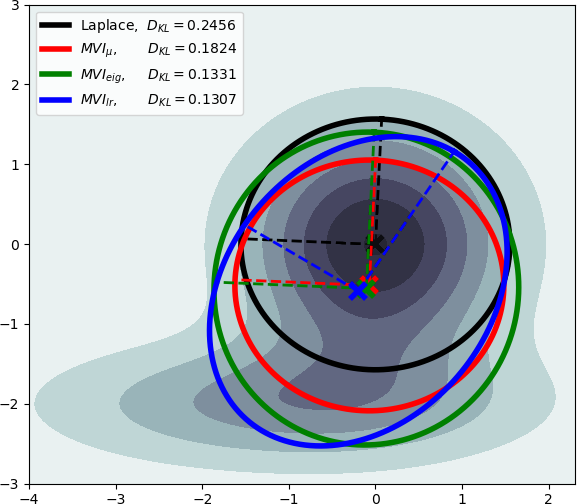}}
\caption{Contour plot of $p(\bd{w})$, see Section \ref{sec:2d_posterior}. We also plot the Gaussian Laplace and MVI posteriors with $\pmb{\times}$ for the mean
and an ellipse for the covariance (70\% confidence interval). The dashed lines are the axes of each ellipse.
The legend reports the $\DKL(q(\bd{w})||p(\bd{w}))$ in each case. The figure is viewed best in digital form.}
\label{fig:2d_example}
\end{figure}

%#####################################################################################################
\section{Applications}
\label{sec:applications}
%#####################################################################################################

We first demonstrate  the behaviour of the proposed MVI posteriors on two synthetic examples.
We then proceed with experiments on benchmark problems.

%-------------------------------------------------------------------------------------------------------------------
\subsection{Illustration with 2D posterior}
\label{sec:2d_posterior}
%-------------------------------------------------------------------------------------------------------------------

We illustrate the MVI posteriors on a synthetic 2D  example where we specify the true, target posterior as a mixture of two Gaussian components:
\begin{align}
p({\bd{w}}) &= \frac{2}{3}\mathcal{N}(\bd{w} | \begin{bmatrix} 0 \\ 0 \end{bmatrix},\bd{I}_2) + \frac{1}{3}\mathcal{N}(\bd{w} | \begin{bmatrix} -1.0 \\ -2.0 \end{bmatrix}, \begin{bmatrix}
3.5 & 0\\ 
0 & 0.3
\end{bmatrix}) \ .
\end{align}

Fig.~\ref{fig:2d_example} displays a contour plot of $p(\bd{w})$.
We approximate $p(\bd{w})$ with LA and the three proposed MVI posteriors and plot them in Fig.~\ref{fig:2d_example}.
The figure also reports the $\DKL(q(\bd{w})||p(\bd{w}))$ of each approximating posterior to the true posterior.
Here $\DKL(q(\bd{w})||p(\bd{w}))$ is calculated numerically as there is no, at least not straightforward,  closed-form expression for it.
We observe that LA (black), by design, places its mean on the mode of $p(\bd{w})$.
All other approximations place their means on alternative locations, but fairly close to one another. 
Even \MVImu\ (red), whose covariance is constrained to be equal to that of LA, can shift its mean to a better location so that it covers more of the target density.
We also note how the axes of \MVIeig\ (green) are parallel by design, but scaled compared to the axes of  \MVImu\ (red), i.e.~we see that the red ellipse of \MVImu\ is contained in the green ellipse of \MVIeig. By scaling its axes,  \MVIeig\ achieves a lower $\DKL$.
\MVIlr\ has the flexibility of rotating its covariance and, in this case, achieves the lowest $\DKL$ to the true posterior $p(\bd{w})$.

%-------------------------------------------------------------------------------------------------------------------
\subsection{Robust regression on synthetic task}
\label{sec:cauchy}
%-------------------------------------------------------------------------------------------------------------------

 \begin{table}[!t]
\renewcommand{\arraystretch}{1.3}
\caption{LPD (higher is better) and MSE (lower is better) on test data for Cauchy regression over $100$ runs.}
\label{table_example}
\centering
\begin{tabular}{l|c|c|c|c|c}
\hline
\bfseries                & \bfseries Laplace       & \bfseries \MVImu             & \bfseries \MVIeig               & \bfseries \MVIlr       & \bfseries \VIdiag       \\
\hline\hline
LPD                       &              -0.818            &           -0.771                    & $\mathbf{ -0.722}$           &    -0.726                  &   -0.736 \\
MSE                      &                0.155          &             0.142                     &             0.129                   &    $\mathbf{0.127}$ &     0.134\\
\hline
\end{tabular}
\label{tbl:cauchy_regression}
\end{table}

\begin{figure*}[t!]
\centerline{\includegraphics[scale=0.47]{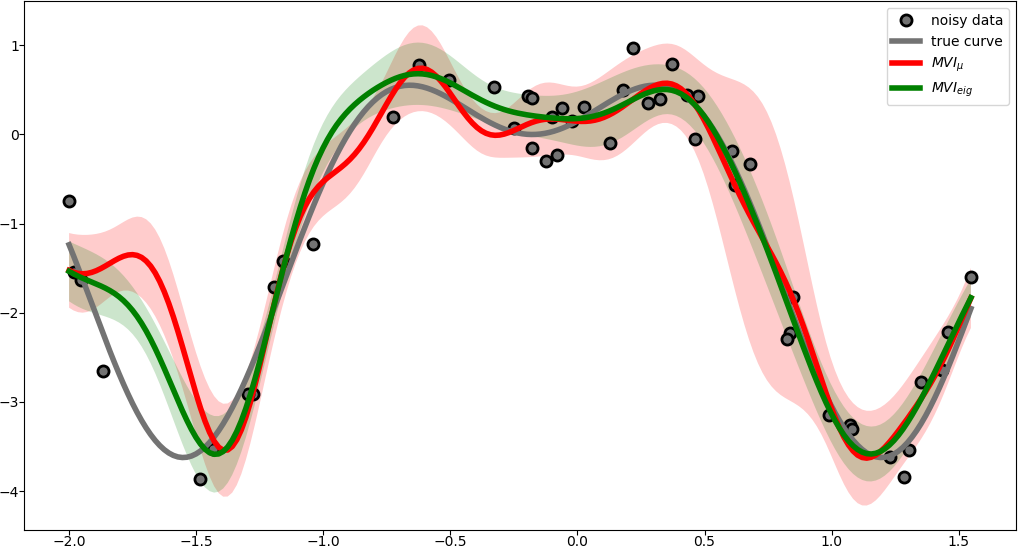}}
\caption{Regression task  in Section \ref{sec:cauchy}, on data corrupted with uniform noise. 
The mean predictions by \MVImu\ (red) and \MVIeig\ (green) are plotted as solid lines. The shaded region around the means corresponds to $\pm$ 2 standard deviations of the predictions.
Predictions are obtained through $S^\prime$ number of samples drawn from the corresponding approximating posterior  $q(\bd{w})$.}
\label{fig:cauchy_regression}
\end{figure*}

We experiment with a regression task with  $N=50$ input-target pairs $(x_n,y_n)$, 
$x_n,\ y_n\in\RR{}$. 
Inputs $x_n$ are drawn uniformly in $[-10.0, +10.0]$.
Targets $y_n$ are generated through the expression 
\begin{equation}
y_n = 0.3x_n\sin(0.7x_n) - 0.03x_n^2
\label{eq:curve}
\end{equation}
and corrupted with i.i.d.~noise drawn from the uniform distribution with support $[-0.5,+0.5]$.
This is a regression task where adopting a Gaussian likelihood would lead to poor results
as it cannot adequately explain the noise. We adopt a Cauchy density instead. 
The unnormalised log-posterior reads:
\begin{equation}
\log \prod_{n=1}^N f(y_n; \bd{w}^T\bd{\phi}_n, \gamma) + \ln \mathcal{N}(\bd{w} | \bd{0},\alpha^{-1}\bd{I}_D) \ ,
\end{equation}
where $f(y; \mu, \gamma)= \left( \pi \gamma \  [1 + (\frac{y-\mu}{\gamma})^2 ]  \right)^{-1}$
is the Cauchy density. Additionally, we  have  calculated a set of $M$ radial basis functions on the data inputs:
\begin{equation}
\bd{\phi}_n = [ \phi(\bd{x}_n;r,\bd{c}_1), \ \dots, \ \phi(\bd{x}_n;r,\bd{c}_{M}), \ 1]^T \ ,
\label{eq:basis}
\end{equation}
where $\phi(\bd{x}_n;r,\bd{c}_m) = \exp(-\frac{\|\bd{x}_n-\bd{c}_m\|^2}{2r^2})$.
The last element $1$ in \eqref{eq:basis}  serves as a  bias term.
Hence, $\bd{\phi}_n\in\RR{M+1}$ and $\bd{w}\in\RR{D}$ with $D=M+1$.

In this numerical experiment, we generate $100$ datasets with $N=50$ training data items
using \eqref{eq:curve}.  We also  generate $N_{test}=1000$ test data items in precisely the same way.
We report the median log-predictive density (LPD) on test data in Table \ref{tbl:cauchy_regression}.
Best performances are marked with bold. We see that the \MVIeig\ approximation performs the best, hence we mark it in bold.
When looking at the results, we established that \MVIeig\ performs statistically significantly better than  LA, \MVImu\ and \VIdiag .
However, as \MVIeig\  does not outperform \MVIlr\  with statistical significance, we do not mark it additionally with $\bullet$ marker.
In terms of error rate, we see that \MVIlr\ performs best and marginally better than \MVIeig . 
Finally, in figure \ref{fig:cauchy_regression} we show the true underlying curve, specified in \eqref{eq:curve},
the observed training data as filled circles along with the regressions induced by \MVImu\ and \MVIeig.

%

%-------------------------------------------------------------------------------------------------------------------
\subsection{Logistic Regression}
\label{sec:logistic_regression}
%-------------------------------------------------------------------------------------------------------------------

 \begin{table*}[!t]
\renewcommand{\arraystretch}{1.3}
\caption{Median LPD on test data for logistic regression over $100$ runs on the datasets (higher is better).}
\label{table_example}
\centering
\begin{tabular}{l|c|c|c|c|c|c|c|c}
\hline
\bfseries  Dataset & Q &$N$ &$N_{test}$ & \bfseries Laplace  & \bfseries \MVImu & \bfseries \MVIeig & \bfseries \MVIlr & \bfseries \VIdiag \\
\hline\hline
Banana                 &2 &      400      &     4900      & -1238.76               &  -1219.19                     & -1221.41                   & $\mathbf{-1212.19}^\bullet$ & -1253.36 \\
Breast cancer       &9&      200      &         77      &     -42.82               &  -42.65                         &    -42.53                    &  $\mathbf{-42.38}$                &     -45.42  \\
Diabetis                &8&       468     &       300      &  -145.98               & -145.468                      &  -145.31                   & $\mathbf{-144.89}^\bullet$  & -193.27 \\
Solar                    &9&       666     &       400      &  -232.64              &  -232.37                     &   -232.42                   & $\mathbf{-232.07}^\bullet$  & -234.52 \\
German               &20&      700    &        300      &  -151.71               & -151.42                      & -151.31                     & $\mathbf{-150.70}^\bullet$  & -179.29 \\
Heart                   &13&      170    &        100      &    -39.25              & -38.973                       & -38.96                       & $\mathbf{-38.62}$               & -48.37 \\
Image                  &18&    1300    &      1010      &  -304.33               & -291.91                       & -284.19                    & -284.60                               & $\mathbf{-283.80}$ \\
Ringnorm            &20&      400    &     7000      &  -309.87                & $\mathbf{-308.80}$    & -319.67                    & -309.80                               & -342.627 \\
Splice                  &60&    1000   &     2175      & -1156.80                 & -900.00                      & $\mathbf{-897.33}$  & -900.30                              & -899.501 \\
Thyroid                & 5&      140    &         75      & -11.01                  & -10.189                        & -10.189                       & $\mathbf{-9.844}^\bullet$    & -10.280 \\
Titanic                  & 3&      150   &     2051      & $\mathbf{-1018.92}^\bullet$  & -1019.59                      & -1021.7                       & -1020.62                             & -1023.91 \\
Twonorm              &20&     400   &    7000      &-452.57                & -450.716                      & -461.10                    & $\mathbf{-447.28}$             &-543.115\\
Waveform           &21&      400  &     4600      & -947.66                & $\mathbf{-946.49}$   & -948.62                      & -950.31                               & -969.61 \\
\hline
\end{tabular}
\label{tbl:logistic_regression_lpd}
\end{table*}

 \begin{table}[!t]
\renewcommand{\arraystretch}{1.3}
\caption{Median error rate $\%$ on test data for logistic regression over $100$  runs on the datasets (lower is better).}
\label{table_example}
\centering
\begin{tabular}{l|c|c|c|c|c}
\hline
\bfseries  Dataset  & \bfseries Laplace  & \bfseries \MVImu & \bfseries \MVIeig & \bfseries \MVIlr & \bfseries \VIdiag \\
\hline\hline
Banana                 & 11.76 & 11.49 & 11.53 & $\mathbf{11.47}$  & 11.74 \\
Breast cancer       & 28.95 & 28.84 & 28.84 & $\mathbf{28.71}$ & 28.98 \\     
Diabetis                & 24.79 & 24.65 & 24.67 & $\mathbf{24.38}$ & 34.33 \\       
Solar                     & 35.15 & 35.14 & 35.16 & $\mathbf{35.03}$ & 35.62 \\             
German                & 26.06 & 25.91 & 25.96 & $\mathbf{25.75}$ & 29.00 \\          
Heart                    & 18.05 & 17.81 & 17.68 & $\mathbf{17.45}$ & 23.04 \\
Image                   & 13.96 & 13.38 & 12.86 & 13.14 & $\mathbf{12.86}$ \\      
Ringnorm             &   1.95 & $\mathbf{1.87}$ & 1.90 & 1.91 & 1.93 \\
Splice                   & 25.74 & 18.98 & 18.96 & 18.89 & $\mathbf{18.71}$ \\                  
Thyroid                 & 6.35  & 6.08   & 6.09    & $\mathbf{6.00}$  & 6.13 \\
Titanic                  & $\mathbf{23.31}$ & 23.31 & 23.33 & 23.32 & 23.62 \\
Twonorm              & 2.92 & 2.86 & 2.92 & $\mathbf{2.84}$ & 2.96  \\
Waveform           & 10.40 & $\mathbf{10.33}$ & 10.36 & 10.34 & 10.44 \\
\hline
\end{tabular}
\label{tbl:logistic_regression_error}
\end{table}

We experiment with logistic regression \cite[Chapter $4$]{Bishop}.
The data are $N$ input-label pairs $(\bd{x}_n,y_n)$ with $\bd{x}_n\in\RR{Q},\ y_n\in\{0,1\}$.
Just like in Section \ref{sec:cauchy},  equation \eqref{eq:basis}, we calculate a set of radial basis functions $\bd{\phi}_n\in\RR{M+1}$ on the data inputs  $\bd{x}_n$.
%where the last $(M+1)-$th element is set to $1$.
The weights are given by $\bd{w}\in\RR{D}$ with $D=M+1$. The unnormalised log-posterior reads:
\begin{align}
\ln \prod_{n=1}^N \sigma(\bd{\phi}_n^T \bd{w})^{y_n}  (1-\sigma(\bd{\phi}_n^T \bd{w}))^{1-y_n}  + \ln \mathcal{N}(\bd{w} | \bd{0},\alpha^{-1}\bd{I}_D) \ .
\label{eq:logistic_regression}                                           
\end{align}

To avoid inadvertently selecting single datasets on which the proposed algorithm performs well, %\footnote{i.e.~``cherry-picking".}, 
we experiment with the entire collection of datasets 
 preprocessed by R\"atsch et al\footnote{http://www.raetschlab.org/Members/raetsch/benchmark}. 
Each dataset has been standardised and split into $100$ training and testing instances, except for \textit{Image} and \textit{Splice} that have 20 splits.
 We approximate the log-posterior in \eqref{eq:logistic_regression} with LA, the MVI posteriors and \VIdiag.
To initialise the hyperparameters $\bd{\theta}=(M,r,\alpha, \bd{c}_1,\dots, \bd{c}_{M})$ in Laplace, we proceed as follows:
per dataset, we run LA for  $10$ iterations for each combination of $M\in\{10,20,30\}$ and $10$ randomly drawn pairs
 $r\sim\mbox{Uniform}(0,1)$, $\alpha\sim\mbox{Uniform}(0,1)$, i.e.~a total of $30$ combinations. 
 The centres $\bd{c}_m$ are determined by K-means for each choice of $M$.
 The combination with the highest lower bound (we are maximising) is declared the winner and
used to initialise LA which is then run for a maximum of $1000$ iterations.
The MVI posteriors and hyperparameters in the respective objectives are initialised using the optimised Laplace
posterior and hyperparameters, as described in Section \ref{sec:initialisation}.

We report the median log-predictive density (LPD) on test data in Table \ref{tbl:logistic_regression_lpd}, along with details about the datasets. Best performances are marked with bold. Best performances that differ in a statistically significant way to all other performances (see Section \ref{sec:measuring_performance}) are additionally marked with a $\bullet$ marker.
Table \ref{tbl:logistic_regression_lpd} reveals that, in general, the proposed MVI posteriors perform better than LA or \VIdiag. 
In particular, we see that \MVIlr\ scores better on a number of datasets and that the difference in performance is often statistically significant.
Table \ref{tbl:logistic_regression_error} displays the results on error rates. We see that all methods achieved more or less the same error rates with no performance being statistically significantly superior. Nonetheless, we do observe a few exceptions, e.g.~on datasets \textit{Splice} and \textit{Diabetis} LA and \VIdiag\ respectively perform noticeably poorer.

%-------------------------------------------------------------------------------------------------------------------
\subsection{Multiclass Logistic Regression}
\label{sec:multiclass_regression}
%-------------------------------------------------------------------------------------------------------------------

 \begin{table*}[t!]
\renewcommand{\arraystretch}{1.3}
\caption{Median LPD on test data for multiclass logistic regression over $100$  runs on the datasets (higher is better).}
\label{table_example}
\centering
\begin{tabular}{l|c|c|c|c|c|c|c|c|c}
\hline
\bfseries  Dataset  & K & Q &$N$ &$N_{test}$&\bfseries Laplace       & \bfseries \MVImu             & \bfseries \MVIeig               & \bfseries \MVIlr                & \bfseries \VIdiag \\
\hline\hline
Ecoli                      & 8 & 7  &236&100&         -50.32          &               -48.80                   &   -49.31                               & $\mathbf{-48.52}^\bullet$ &                  -51.39\\
Crabs                    & 4 & 5  &140&  60&         -64.59           &               -64.11                     & -64.28                               & $\mathbf{-64.10}$            &                   -68.92\\
Iris                         & 3 & 4  &105&  45&           -9.06           &                -7.53                     &   $\mathbf{-6.42}$              &                  -7.46              &                    -8.17 \\
Soybean                & 4 & 35&  33&  14&           -4.10           &                -2.35                    &   $\mathbf{-0.66}^\bullet$   &                 -2.36                &                    -1.67\\
Wine                      & 3 & 13&125&  53&           -5.72           &                -4.01                    &   $\mathbf{-3.33}^\bullet$   &                  -3.94               &                    -4.66 \\
Glass                     & 6 &  9 & 150& 64&           -61.35         &              -60.39                    &   $\mathbf{-59.79}$            &               -60.44               &                   -76.26 \\
Vehicle                   & 4 & 18& 593&293&     -159.783         & $\mathbf{-158.39}^\bullet$   &           -158.60                    &             -159.15                &                 -174.725\\
Balance                 &  3 &  4 & 438&187&      -23.5734        &  $\mathbf{-22.7321}$           &            -23.2197                 &               -23.078               &  -31.587              \\ 
\hline
\end{tabular}
\label{tbl:multiclass_lpd}
\end{table*}

 \begin{table}[t!]
\renewcommand{\arraystretch}{1.3}
\caption{Median error rate $\%$ on test data for multiclass logistic regression over $100$  runs on the datasets (lower is better).}
\label{table_example}
\centering
\begin{tabular}{l|c|c|c|c|c}
\hline
\bfseries  Dataset  & \bfseries Laplace       & \bfseries \MVImu             & \bfseries \MVIeig               & \bfseries \MVIlr                & \bfseries \VIdiag \\
\hline\hline
Ecoli                      &         17.22                  &            16.75                    &                   17.23              &          $\mathbf{16.73}$    &            17.87        \\
Crabs                    &         55.75                  &         $\mathbf{54.98}$     &                    55.10             &                   55.00              &            58.96        \\
Iris                         &           9.24                  &          $\mathbf{7.57}$     &                      7.86             &                     7.58              &              9.76        \\
Soybean               &          13.12                 &             4.83                     &  $\mathbf{3.85}^\bullet$    &                    4.86              &              9.33         \\
Wine                     &            4.97                 &           $\mathbf{3.11}$     &                      3.31             &                    3.14               &             4.72          \\
Glass                    &          42.93                 &                41.14                &       $\mathbf{39.79} $       &                   41.14              &            53.01         \\
Vehicle                 &           32.60                &           $\mathbf{32.15}$   &                      32.34           &                    32.22               &          34.61        \\
Balance                &           6.56                  &           $\mathbf{6.07}$     &                     6.27               &                    6.25               &            8.06         \\ 
\hline
\end{tabular}
\label{tbl:multiclass_error_rate}
\end{table}

Similarly to logistic regression, multiclass logistic regression \cite[Chapter $4$]{Bishop} does not allow 
direct Bayesian inference as the use of the softmax function renders integrals over the likelihood term intractable. 
The unnormalised log-posterior reads:
\begin{align}
 \ln \prod_{n=1}^N \prod_{k=1}^K p(C_k|\bd{\phi}_n)^{y_{nk}} + \ln \prod_{k=1}^K \mathcal{N}(\bd{w}_k | \bd{0},\alpha^{-1}\bd{I}_D) \ ,
\end{align}
where $K$ denotes the total number of classes.
The data are input-label pairs $(\bd{x}_n,\bd{y}_n)$ with $\bd{x}_n\in\RR{Q}$.
Vectors $\bd{y}_n$ are binary vectors encoding class labels using a $1$-of-$K$ coding scheme, e.g. $[0\ 1\ 0]$ encodes class label $2$ in a $3$-class problem.
The probability $p(C_k|\bd{\phi}_n)$ of the $n$-th data item belonging to class $C_k$ is modelled via the softmax function:
\begin{equation}
p(C_k|\bd{\phi}_n)=\frac{\exp(\bd{\phi}_n^T\bd{w}_k)}{\sum_{\ell=1}^K \exp(\bd{\phi}_n^T\bd{w}_\ell)} \ ,
\end{equation}
where each class $C_k$ is associated with a weight vector $\bd{w}_k\in\RR{D}$, with $D=M+1$.
The basis functions $\bd{\phi}_n\in\RR{M+1}$ are defined in the same way as in Section \ref{sec:cauchy}.
We initialise hyperparameters  $\bd{\theta}=(M,r,\alpha, \bd{c}_1,\dots, \bd{c}_{M})$ in the same way as described in Section \ref{sec:logistic_regression}.

% \begin{table*}[!t]
%\renewcommand{\arraystretch}{1.3}
%\caption{Median LPD on test data for robust GP over $100$  runs on the datasets (higher is better).}
%\label{table_example}
%\centering
%\begin{tabular}{l|c|c|c|c|c|c|c}
%\hline
%\bfseries  Dataset  & Q  & $N$ & $N_{test}$ & \bfseries Laplace       & \bfseries \MVImu             & \bfseries \MVIeig               & \bfseries \MVIlr		    \\
%\hline\hline
%Boston housing     & 13 &355  &    151        & -0.35                          & -0.33                                & $\mathbf{-0.21} ^\bullet$    & -0.33                                \\
%Concrete strength &   8 &721  &    309        & -0.35                          & -0.34                                &   -0.34                                 &  $\mathbf{-0.26}^\bullet$  \\
%\hline
%\end{tabular}
%\label{tbl:robust_gp_lpd}
%\end{table*}

To avoid inadvertently selecting single datasets on which the proposed algorithm performs well,
we experiment with the collection of multiclass datasets used in the work of \cite{Psorakis2010} in a different context.
Details of the datasets are shown in Table  \ref{tbl:multiclass_lpd}.
We standardise the data column-wise to zero mean and unit standard deviation.
Using random subsampling, we split each dataset $100$  times into a training ($70\%$ of the data) and testing ($30\%$) set.
We report the median LPD for each dataset and algorithm in Table \ref{tbl:multiclass_lpd} and median error rate in Table \ref{tbl:multiclass_error_rate}.
Again, best performances are marked in bold. We mark the best performance with a $\bullet$ marker if it is found to be statistically significant using the same
two conditions described in Section \ref{sec:measuring_performance}.
The results show an improvement over LA and the use of a factorised posterior in \VIdiag. 
We also see that \MVIeig\ performs well on this set of problems in terms of LPD, though the picture is not as clear in terms of error rate in Table \ref{tbl:multiclass_error_rate}.
Finally, we note the low performance of all methods on the dataset \textit {Crab}, evidently in Table \ref{tbl:multiclass_error_rate}.
This may be perhaps attributed to the particular choice of the RBF kernel made here, though other kernels (cf  \cite{Psorakis2010}) may be more appropriate.

\section{Discussion and Conclusion}
%#####################################################################################################

We proposed Mixed Variational Inference (MVI) as a method for approximate Bayesian inference in non-conjugate models.
MVI makes use of the posterior obtained via the Laplace approximation and the objective function provided by variational inference.
The adoption of the Laplace posterior helps with capturing a-posteriori correlations; 
the partial adaptation of the Laplace posterior, in the form of the proposed MVI posteriors, helps limit the number of free variational parameters 
that need to be optimised.
The numerical results show that the MVI posteriors have the potential to improve on the performance of the Laplace approximation 
and on the performance of the commonly adopted factorised Gaussian posterior in variational inference.

Strictly speaking, however, one should be aware of the fact that a posterior $q(\bd{w})$ that approximates the true posterior better, does not necessarily guarantee improved log-predictive density;
vice versa, a ``naive'' approximating posterior  (e.g.~factorised) may in principle provide satisfactory predictive performance.
This observation has been previously stated in \cite{Nickisch2008} where a variety of approximations are evaluated in the context of Gaussian process binary classification. Therein it is stated that, in principle, even a poor approximation in terms of posterior moments can still provide good predictions.
After all, as far as variational approximations are concerned, it is evident in  objective  \eqref{eq:objective_VI} that the goal is to find a $q(\bd{w})$ that is as close as possible to the true posterior; this does not necessarily correlate with improved predictive performance.
Nevertheless, one does expect in practice that an approximation that captures posterior correlations in the parameters to be more useful than
a factorised approximation that practically draws the parameters independently of one another when making predictions (see \eqref{eq:mll}).
But beyond this expectation, it is admittedly difficult to anticipate what approximation may perform best.
Indeed, in the numerical experiments we notice that while \MVIlr\  seems well suited for logistic regression (see Table \ref{tbl:logistic_regression_lpd}),
this is not necessarily the case in multiclass logistic regression (see Table \ref{tbl:multiclass_lpd}).

In its present form, MVI is limited to Gaussian posteriors. 
It would be interesting to extend MVI to non-Gaussian posteriors, though at first sight it seems that its dependence on the Laplace approximation considerably limits it. An interesting direction, inspired by \cite{Gershman2012}, would be to postulate a posterior $q(\bd{w})$ based on a mixture of Gaussians, where the covariance of each Gaussian  comes from a Laplace approximation performed at a different mode.
Forming an approximating posterior using multiple modes procured by the Laplace approximation has  been previously suggested in \cite{gelman2013bayesian}.
However, therein no objective function akin to \eqref{eq:objective_SVI} is guiding the inference of the posterior. This could be potentially addressed by extending MVI
so that $q(\bd{w})$ is now a Gaussian mixture whose covariance matrices are given by the Laplace approximation and partially updated as suggested in Sections \ref{sec:mvi_mu}, \ref{sec:mvi_eig}, \ref{sec:mvi_lr}.
We reserve such investigations for future research.

%-------------------------------------------------------------------------------------------------------------------
\section*{Acknowledgment}
%-------------------------------------------------------------------------------------------------------------------

The author acknowledges the useful discussions and encouragement of Christoph Schn\"orr, Kai Polsterer and Ata Kaban.

\IEEEtriggeratref{18}
\bibliographystyle{IEEEtran}
\bibliography{biblio}

\end{document}